\documentclass{article}
\usepackage{spconf,amsmath,graphicx}
\usepackage{amssymb}
\usepackage{booktabs}

\usepackage[pagebackref,breaklinks,colorlinks]{hyperref}
\usepackage{makecell}
\usepackage{setspace}

\usepackage[capitalize]{cleveref}
\crefname{section}{Sec.}{Secs.}
\Crefname{section}{Section}{Sections}
\Crefname{table}{Table}{Tables}
\crefname{table}{Tab.}{Tabs.}

\title{RACon: Retrieval-Augmented Simulated Character Locomotion Control}
\name{Yuxuan Mu\textsuperscript{1}, Shihao Zou\textsuperscript{1}, Kangning Yin\textsuperscript{2}, Zheng Tian\textsuperscript{2}\sthanks{Corresponding author: tianzheng@shanghaitech.edu.cn}, Li Cheng\textsuperscript{1}, Weinan Zhang\textsuperscript{3}, Jun Wang\textsuperscript{4}}
\address{\textsuperscript{1}Department of Electrical and Computer Engineering, University of Alberta\\
\textsuperscript{2}School of Creativity and Art, ShanghaiTech University\\
\textsuperscript{3}Department of Computer Science \& Engineering, Shanghai Jiao Tong University\\
\textsuperscript{4}Department of Computer Science, University of London College \thanks{This work is supported by Shanghai Sailing Program (23YF1427600). The SJTU team is partially supported by Shanghai Municipal Science and Technology Major Project (2021SHZDZX0102) and National Natural Science Foundation of China (62322603, 62076161).}}
\begin{document}
%
\maketitle
\begin{abstract}
In computer animation, driving a simulated character with lifelike motion is challenging. Current generative models, though able to generalize to diverse motions, often pose challenges to the responsiveness of end-user control. To address these issues, we introduce \textbf{RACon}: \textbf{R}etrieval-\textbf{A}ugmented Simulated Character Locomotion \textbf{Con}trol. Our end-to-end hierarchical reinforcement learning method utilizes a retriever and a motion controller. The retriever searches motion experts from a user-specified database in a task-oriented fashion, which boosts the responsiveness to the user's control. The selected motion experts and the manipulation signal are then transferred to the controller to drive the simulated character. In addition, a retrieval-augmented discriminator is designed to stabilize the training process. Our method surpasses existing techniques in both quality and quantity in locomotion control, as demonstrated in our empirical study. Moreover, by switching extensive databases for retrieval, it can adapt to distinctive motion types at run time. We will release our code upon acceptance.
\end{abstract}
\begin{keywords}
Locomotion control, physical simulation, reinforcement learning, retrieval augmented model
\end{keywords}
\vspace{-2mm}

\section{Introduction}
\label{sec:intro}

Recent advancements in the simulation of virtual character motion can be largely credited to the advent of physics-based deep reinforcement learning methods, such as those outlined in~\cite{2022adversarialmotion}. Despite these advancements, it remains a considerable challenge to create systems that allow end users to conveniently manipulate character locomotion while simultaneously ensuring the motions generated are convincingly natural and realistic. Moreover, the system should be capable of seamlessly transitioning to new motion types in real-time.

\begin{figure}[t]
    \centering
    \includegraphics[width=0.9\columnwidth]{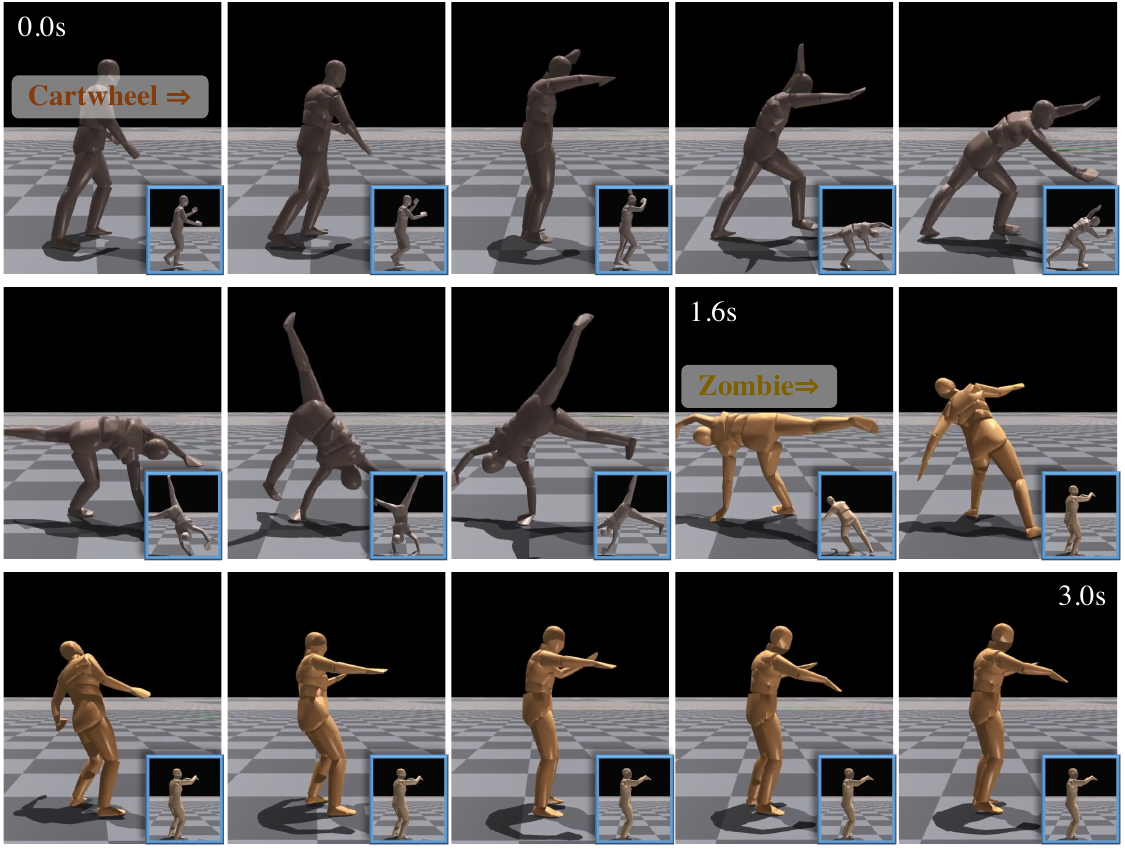}
    \vspace{-2mm}
    \caption{\textbf{Example of switch between two distinct motion types, cartwheel and zombie walk, as the response of user's switching databases.} The retrieved reference motion is shown in the bottom right. The color of the character marks the style-distinctive expert retrieval database being used at test time. Brown: Cartwheel. Yellow: Zombie. Note that zombie is a new motion type added by the user at test time which is not involved in training. Our results show a natural transition whenever the retrieval database changes from one to another.
    }
    \vspace{-4mm}
    \label{fig:transcart2zomb}
\end{figure}

Motion tracking has been utilized by several recent studies~\cite{liu2016guided} to imitate a library of pre-existing 3D expert motion sequences. Often a motion planner is employed to synchronize expert motions with the character simulation over time, relying on manually constructed pose features~\cite{bergamin2019drecon}. Although these methods have been demonstrated to generate high-fidelity motions, their performance largely depends on the quality and quantity of the available expert motions. These are typically carefully curated and often limited in scope, resulting in mediocre generalization when faced with complex scenarios or large datasets. Their effectiveness diminishes further when dealing with novel motion types or styles \cite{liu2016guided, peng2018deepmimic}.
In response to these limitations, recent research has considered stochastic generative models such as GANs~\cite{peng2021amp,peng2022ase,2022adversarialmotion} and VAEs~\cite{yao2022controlvae, won2022physics, ling2020character}. These models have produced impressive results, demonstrating their ability to generalize to unseen motion types or scenarios. However, as a black-box framework, these models pose a challenge for end-users who wish to explicitly determine or edit specific poses or skills~\cite{peng2021amp, yao2022controlvae}.

The recent successes of Retrieval-Augmented Models (RAM) in Natural Language Processing, showing case its straightforward knowledge acquisition and scalability, has inspired us to employ innovative retrieval and RAM techniques in our motion control context. Our work aims to enhance task responsiveness, generate natural and realistic animations, and alleviate the need to fine-tune a complex controller. Consequently, we propose an end-to-end retrieval-augmented (RA-) solution that utilizes hierarchical reinforcement learning (HRL) for task-oriented learnable retrieval and physics-based character control. 

The workflow of our approach is as follows: The direction and velocity specified by the gamepad stick are combined with the current character state and fed into a retrieval policy. This policy predicts an adaptive query, which then searches for the reference motion from an expert motion database. The controller receives this reference and the goal is to generate driven signals for the character's actuators within the simulator. The retrieved and simulated motions are subsequently filtered through a RA-discriminator, adhering to the Generative Adversarial Imitation Learning (GAIL) strategy~\cite{ho2016generative}. This step serves as a motion prior, ensuring the coherence of retrieved motions and the naturalness of simulated ones.

Components such as the goal, prior, and simulated state feedback are fed back to both the retrieval and control modules. Within this context, two reinforcement learning frameworks, task-oriented retrieval and embodied agent manipulation, are seamlessly integrated into our approach. Thanks to our RA-HRL design, our system can generate high-fidelity character motions and adapt to distinctive motion types without extensive professional tuning or selection. Meanwhile, it remains interpretable, providing evidence of retrieved reference motion from user-selected Motion Capture (MoCap) databases (as depicted in Fig.~\ref{fig:transcart2zomb}).

Our contributions are summarized below:
\begin{itemize}
    \item An end-to-end integrated approach is proposed, which incorporates two RL frameworks as an HRL system, and a RA-discriminator as a training-time motion prior. A set of rewards is designed to jointly optimize the two policies at both the policy level and the system level. This leads to stable run-time performance and alleviates the mode collapse issue typically found in GAIL models, scaling to large-scale datasets.
    \item Our approach is demonstrated to drive character locomotions of both high fidelity and diversity, outperforming the state-of-the-art methods both quantitatively and qualitatively. It is additionally capable of transiting to different motion types in-situ, in responding to end-user choices.
\end{itemize}



\section{Our Approach}

\begin{figure}[ht] 
    \centering
    \includegraphics[width=\columnwidth]{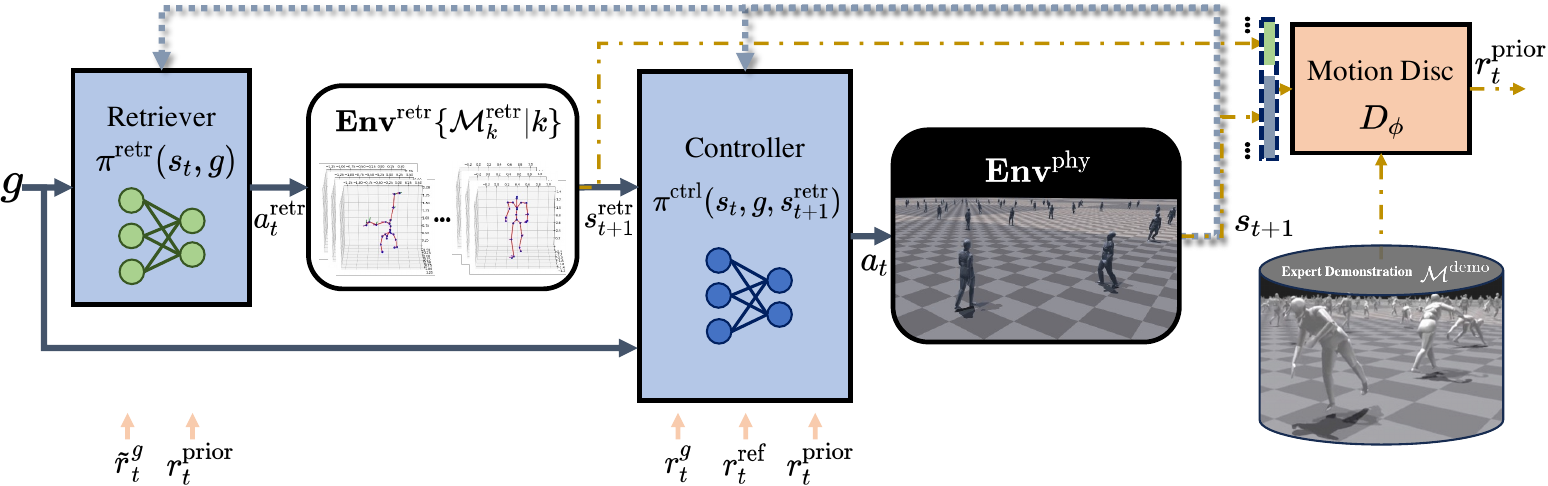}
    \caption{\textbf{The pipeline of our proposed RACon HRL System.} The forward process is shown in solid arrows, and the state feedback process is shown in dot arrows. Our system incorporates two frameworks, $\langle \pi^{\text{retr}}, \text{Env}^{\text{retr}}\rangle$ and $\langle \pi^{\text{ctrl}}, \text{Env}^{\text{phy}}\rangle$, working in the 'manager-worker' fashion. These two frameworks share the same goal signal $g$ and feedback state $s_{t+1}$ and are trained in an end-to-end manner with policy-level rewards (\textit{i.e.} $\tilde{r}^{\text{g}}_{t}$, $r^{\text{ref}}_{t}$) and system-level rewards (\textit{i.e.} $r^{\text{g}}_{t}$, $r^{\text{prior}}_{t}$). The optimization details of the retriever and the controller are elaborated in \cref{sec:MER} and \cref{sec:ReferenceGuided} respectively. In addition, the yellow dash arrows mark the workflow of retrieval-augmented motion discriminator described in~\cref{sec:motionprior} which provides a robust motion prior reward for both policies.}
    \vspace{-1mm}
    \label{fig:pipline}
\end{figure}

Our method, depicted in Fig.\ref{fig:pipline}, incorporates two frameworks: task-oriented retrieval~\cref{sec:MER} and embodied agent manipulation~\cref{sec:ReferenceGuided}. Specifically, we embed the retrieval framework into the embodied agent manipulation framework, acting as a high-level 'manager' that provides 'guidance' to the controller policy. This arrangement operates within a hierarchical reinforcement learning structure, in response to the given goal signal $g$. 
The overarching learning objective is to follow the locomotion control $g$ and mimic human behavior, which is realized through the rewards $r^g$ and $r^\text{pior}$. The prior reward $r^\text{pior}$ is provided by a retrieval-augmented discriminator as detailed in~\cref{sec:motionprior}.
We optimize the system in accordance with the principles of GCRL. 

\subsection{Task-Oriented Learnable Retrieval (TOLR)}
\label{sec:MER}

The idea of motion retrieval revolves around the periodic querying of the database to identify the motion clips that is best compatible in terms of fulfilling user control expectations. The periodically retrieved clips are stitched together to create a complete reference trajectory. Traditional methods like Motion Matching~\cite{bergamin2019drecon} employ hand-crafted features for retrieval, posing a challenge in determining which features are vital for optimal performance whenever there's a modification in the database or the representations (\textit{e.g.}, DoFs).
In contrast, our work introduces an innovative data-driven strategy that employs neural networks to formulate queries based on the current character state and task objectives. By harnessing the learning capabilities of neural networks, our approach can automatically identify significant features for motion retrieval, even within massive and disordered databases.

\begin{figure}[ht]
    \centering
    \includegraphics[width=0.75\columnwidth]{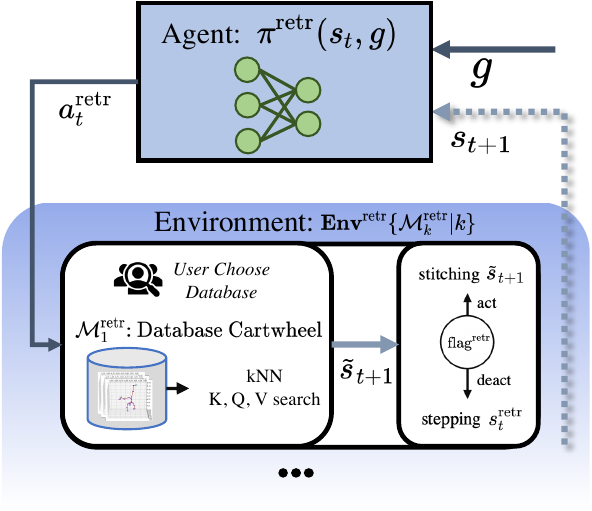}
    \caption{\textbf{The RL framework of Task-Oriented Learnable Retrieval.} 
    Our retrieval environment can hold a set of databases $\{\mathcal{M}^{\text{retr}}_k\}$ of different motion types for the user to choose at run time.
    Given the action $a_t^{\text{retr}}$ as a query $Q$, the environment searches for the most similar key $K$ to index the most suitable value $V$, \textit{i.e.} a motion clip $\tilde s_{t+1}$, from the database. 
    When the $\boldsymbol{\text{flag}^{\text{retr}}}$ is activated, the motion clip will be \textit{stitch}ed to the character by transforming it to align the initial frame's root coordinate to the current character to build $s^\text{retr}_{t+1}$. Otherwise, the $\boldsymbol{\text{flag}^{\text{retr}}}$ is deactivated where the character will continuously \textit{step} along the previous retrieval state $s^\text{retr}_{t \to t+1}$. 
    }
    \vspace{-4mm}
    \label{fig:retrpipline}
\end{figure}

Specifically, TOLR framework illustrated in~\cref{fig:retrpipline} is considered as a goal-augmented Markov decision process. The \textit{environment} of retrieval is defined as the process of kNN-based $K,Q,V$ search where the given query vector $Q$ is employed to search for the most similar key vector $K$ to find the optimized value $V$. These keys are calculated from the motion clips in advance when building the databases at one time, which are constituted as libraries of key-value pairs in the form ${K \colon V | \mathcal{M}^\text{retr}}$. At run time, we first extract and compute the raw query in the same manner as the keys $K$, with features such as initial-frame root velocity, clip average velocity, end effector positions, etc. Then, the retriever policy is used to generate a weight vector, working as the adaptive importance of different features, which adjusts the raw query to be the input of the environment $Q$.

Additionally, the retrieval \textit{environment} can integrate extra plug-in motion databases of preferred types in real-time to perform locomotion in unique motion styles, which allows our system to execute diverse locomotion skills that are tractable for the end-user. 
To achieve this, we train the system by randomly alternating retrieval databases of different motion types. 

\begin{figure}[htb]
    \centering
    \includegraphics[width=0.6\columnwidth]{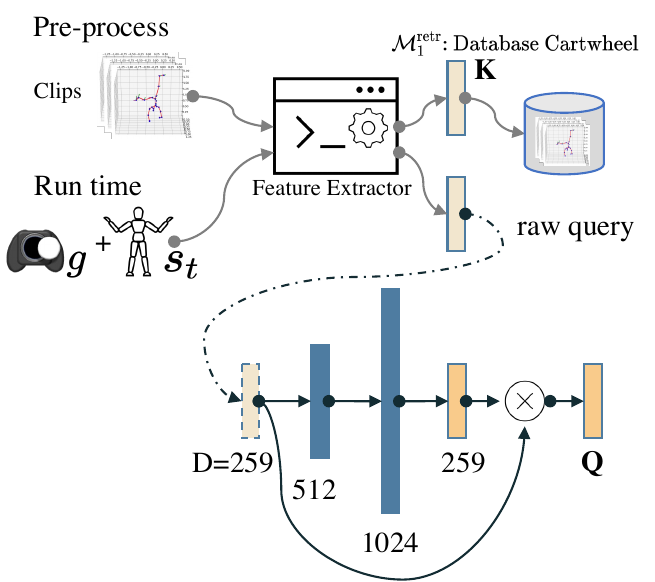}
    \caption{\textbf{The architecture of TOLR.} The non-differentiable feature extractor extracts and computes features from motion clips or current state. The pipeline at the bottom is the actual retrieval policy built by linear layers.}
    \vspace{-2mm}
    \label{fig:retrarch}
\end{figure}

The architecture of our task-oriented learnable retrieval (TOLR) is depicted in~\cref{fig:retrarch}. As the key $K$ is part of the \textit{environment}, which shouldn't be influenced by the agent, modifying the key through an agent policy — following the traditional setting to create the key and query via identical mapping — would conflict with our TOLR RL framework definition. Consequently, we calculate the key in advance when building databases (\textit{i.e.} $\{K \colon V | \mathcal{M}^\text{retr}\}$) at one time, and exclusively use the policy to generate the query at run time. To preserve the essence of classical retrieval, we design the policy to generate adaptive weights to automatically tune the raw query into $Q$ through a Hadamard product, where the raw query is computed in the same manner as the key.

Our optimization objective has a dual focus: maximizing the capability of the method to accomplish control goals, and enhancing the continuity between sequential retrieved motion clips. This is achieved by employing a goal reward and a prior reward, as follows:
\begin{equation}
    {r}^{\text{retr}}_{t}=\tilde{w}^{\text{g}} \tilde{r}^{\text{g}}_{t} + {w}^{\text{prior}} {r}^{\text{prior}}_{t}.
\end{equation}
Here $\tilde{r}^{\text{g}}_{t}=\exp\big(-\mathit{d}\big(g, \tilde{q}_{t}\big)\big)$, where $\mathit{d}(\cdot)$ is the function that evaluates the distance between the current character state and the goal. Specifically, it is the combination of cosine similarity and norm difference of the retrieved trajectory velocity and expected counterpart. $\tilde{q}_t$ is the horizontal root velocity and rotation extracted from the retrieval state $s^\text{retr}_t$. ${r}^{\text{prior}}_{t}$ is determined by a discriminator, as explained in~\cref{sec:motionprior}.

\subsection{Simulated Character Control}
\label{sec:ReferenceGuided}
Although TOLR provides a synchronous reference sub-goal $\hat{g}^{\text{sub}}_t \triangleq {s}^{\text{retr}}_{t+1}$ for the controller $\pi^{ctrl}\left({a}^{\text{ctrl}}_{t}|{s}_{t}, g, {s}^{\text{retr}}_{t+1}\right)$, the learning objective here differs from a mere mimicry term that would require strict adherence to the reference motion and trajectory.
We conceptualize our methodology as an \textit{end effector constraint}, allowing the controller to learn the motion types from specific reference motions, without being trapped by them. Therefore, we incorporate only the end effector feature and root rotations of the reference motion to define the reference reward:
\begin{equation}
    r^{\text{ref}}_{t}=w^{\text{h}}r^{\text{h}}_{t} + w^{\text{rrot}}r^{\text{rrot}}_{t} + w^{\text{ravel}}r^{\text{ravel}}_{t} + w^{\text{local}}r^{\text{loc}}_{t},
\end{equation}
where $r^{\text{h}}_{t}$ represents the root height reward, $r^{\text{rrot}}_{t}$ denotes the root rotation reward independent of yaw, $r^{\text{ravel}}_{t}$ is the root angular velocity reward relative to the root coordinate, and $r^{\text{loc}}_{t}$ measures the error of character body endpoints relative to the root coordinate. More details are described in the supplement.

Our modified mimicry objective allows for some deviation from the original reference, as it doesn't require the character to replicate the rotation at every joint. 
While this flexibility might compromise the naturalness of the motion, it benefits the controller responsiveness by offering a more adaptable curriculum. 
This deviation can then be offset and enhanced through the RA- motion prior ${r}^{\text{prior}}_{t}$, as discussed in \cref{sec:motionprior}. 
Additionally, we incorporate the locomotion control goal into the reward definition for the controller policy:
\begin{equation}
    {r}_{t} = {w}^{\text{g}}{r}^{\text{g}}_{t} + {w}^{\text{ref}}{r}^{\text{ref}}_{t} + {w}^{\text{prior}}{r}^{\text{prior}}_{t},
\end{equation}
where $r^{\text{g}}_{t}=\exp\big(-\mathit{d}\big(g, q_{t}\big)\big)$, similar to $\tilde{r}^{\text{g}}_{t}$, while $q_t$ is from the simulated state $s_t$.

\subsection{Retrieval Augmented Adversarial Motion Prior}
\label{sec:motionprior}

Given a motion dataset, our adversarial framework trains a motion discriminator to predict whether a state transition $\langle s_t,s_{t+1}\rangle$ originates from the real dataset or is a generated sample. Our generated samples can be either simulated state transitions, represented as $(s_t,s_{t+1})\sim \mathcal{M}^{\pi^\text{ctrl}}$, or retrieved state transitions denoted as $(s_t,\tilde{s}_{t+1})\sim \mathcal{M}^{\pi^\text{retr}}$. Here, $\tilde{s}_{t+1}$ is the state retrieved from databases based on the simulated state $s_t$. Our motion discriminator then provides a prior reward, ${r}^{\text{prior}}_{t}$, which fosters coherence for retrieved motions (\cref{sec:MER}) and naturalness for simulated motions (\cref{sec:ReferenceGuided}).
Intriguingly, a composite generated sample can be formed as $(s_{t},s_{t+1},\tilde{s}_{t+2})\sim \mathcal{M}^{\pi}$, as the retrieved anchor state $\tilde{s}_{t+1}$ can be stepped to produce the subsequent $\tilde{s}_{t+2}$, resulting in a retrieval-augmented term of simulated state transition. This integration of generated samples has proven empirically to enhance the system's performance. In what follows, we discuss the effect of this retrieval-augmented design, namely Retrieval Augmented Adversarial Motion Prior.

Previous GAIL-based methods, such as AMP~\cite{peng2021amp}, have shown limitations in terms of stability and efficiency. These methods typically require well-organized, high-quality, and limited-scale datasets, as the generation process might initiate with extremely outlier instances, potentially leading to early-stage model breakdown or unstable learning process. 
To mitigate these issues, our design transforms the framework into a \textit{Conditional GAIL}, augmenting the simulated sample features to $\langle s_{t},s_{t+1},\tilde{s}_{t+2}\rangle$ by utilizing the retrieved state $\tilde{s}_{t+2}$. 
More specifically, the simulated motion and the retrieved expert feature $\tilde{s}_{t+2}$ are concatenated to form a retrieval-augmented simulated sample. In the context of controller training, the RA-term can be seen as a pseudo-label condition drawn from real motion databases and ideally being coherent with the preceding simulated states $\langle s_{t},s_{t+1}\rangle$.
Empirically, one of the most notable characteristics of real motion samples is temporal smoothness. This property facilitates the discriminator's task in distinguishing fake from real samples and guides the synthesis of spatial pose more easily, by considering the consistency between the simulated states and the RA-term.

We define the training objective for the discriminator as: 
\begin{equation}\footnotesize
    \label{eq:discloss}
    \begin{aligned}
        &\mathop{\arg\min}\limits_{\phi}  \mathbb{E}_{\left (s_{t},s_{t+1},{s}_{t+2} \right )\sim \mathcal{M}^{\text{demo}}}\left[ \log D_{\phi}\left(s_{t},s_{t+1},{s}_{t+2}\right)\right]\\ &+\mathbb{E}_{\left (s_{t},s_{t+1},\tilde{s}_{t+2} \right )\sim \mathcal{M}^{\pi}}\left[ \log \left( 1 - D_{\phi}\left(s_{t},s_{t+1},\tilde{s}_{t+2}\right)\right)\right]\\
        &+\frac{w^{\text{gp}}}{2} \mathbb{E}_{\left (s_{t},s_{t+1},{s}_{t+2} \right )\sim \mathcal{M}^{\text{demo}}} \| \nabla_{\phi}D_{\phi}\left(s_{t},s_{t+1},{s}_{t+2}\right)\|^2,
    \end{aligned}
\end{equation}
and the motion prior reward for the two policies as: 
\begin{equation}
    r^{\text{prior}}_{t}=-\alpha\log\max\left[1-D_{\phi}\left(s_{t},s_{t+1},\tilde{s}_{t+2}\right),\epsilon\right],
\end{equation}
where $\alpha$ is a scaling factor to adjust the range of the reward. 
Here, $\mathcal{M}^{\text{demo}}$ represents the demonstration dataset of real motions, while $\mathcal{M}^{\pi}$ denotes the generated dataset. Given that $\tilde{s}_{t+2}$ is retrieved from the real motion database, it can be viewed as the posterior label condition for the simulated transition $\langle s_t,s_{t+1}\rangle$. Accordingly, we can recast the discriminator within the framework of a Conditional GAIL, formalizing it as $D_{\phi}\left(s_{t},s_{t+1}|\tilde{s}_{t+2}\right)$. It's worth noting that the features could extend beyond three continuous states, incorporating $s_{t-i}$ and $\tilde{s}_{t+2+j}$ in implementation. 

\section{Experiment}

\subsection{Motion Database}

Our demonstration dataset of real motions, $\mathcal{M}^{\text{demo}}$, is sourced from AMASS~\cite{mahmood2019amass}, a comprehensive motion dataset that consolidates various MoCap data resources. We extract over 40,000 seconds of expert motion, unified to 30 fps. The databases for retrieval, $\{\mathcal{M}^{\text{retr}}_k\}$, are also constructed from AMASS. We filter common locomotion movements using category labels from BABEL~\cite{punnakkal2021babel}, such as walking, running, turning, and so forth—which amount to approximately 49,000 motion clips, each with a duration of 0.5 seconds. Additionally, we build databases for cartwheel movements ($\sim$6590 clips) and zombie movements (142 clips), respectively. It should be noted that the zombie database is solely employed for testing purposes, and its movement is limited to ``standing still", to simulate a user-given plugin database.

\subsection{Implementation Details}

To ensure fairness and to effectively demonstrate the design of our proposed system, we build $D_{\phi}, \pi^\text{retr}, \pi^\text{ctrl}, V^\text{retr}, V^\text{ctrl}$ with three-layered MLPs of hidden sizes $[1024, 512]$. Both policies are trained with the proximal policy optimization (PPO) method. The discount factor $\gamma$ is set at $0.97$, while the learning rate is established at $5\times10^{-5}$ with Adam as the optimizer. Our simulation is set up on Isaac Gym, with the fundamental system clock at 30 fps. The methods we are comparing are either properly adapted (\textit{i.e.} AMP~\cite{peng2021amp}, ASE~\cite{peng2022ase}) or independently implemented (\textit{i.e.} DReCon~\cite{bergamin2019drecon}) on the same platform and settings.

\subsection{Results}
In the experiment results, we first qualitatively evaluate our method by performing  Cartwheel and Common Locomotion, such as walk and run, and switching motion types at run time. Then, a comprehensive evaluation is conducted on the common locomotion task, compared with the state-of-the-art methods. Lastly, the ablation study for our RA-GAIL and TOLR is presented to further demonstrate the efficacy of our design.

\begin{figure}
    \centering
    \includegraphics[width=0.98\columnwidth]{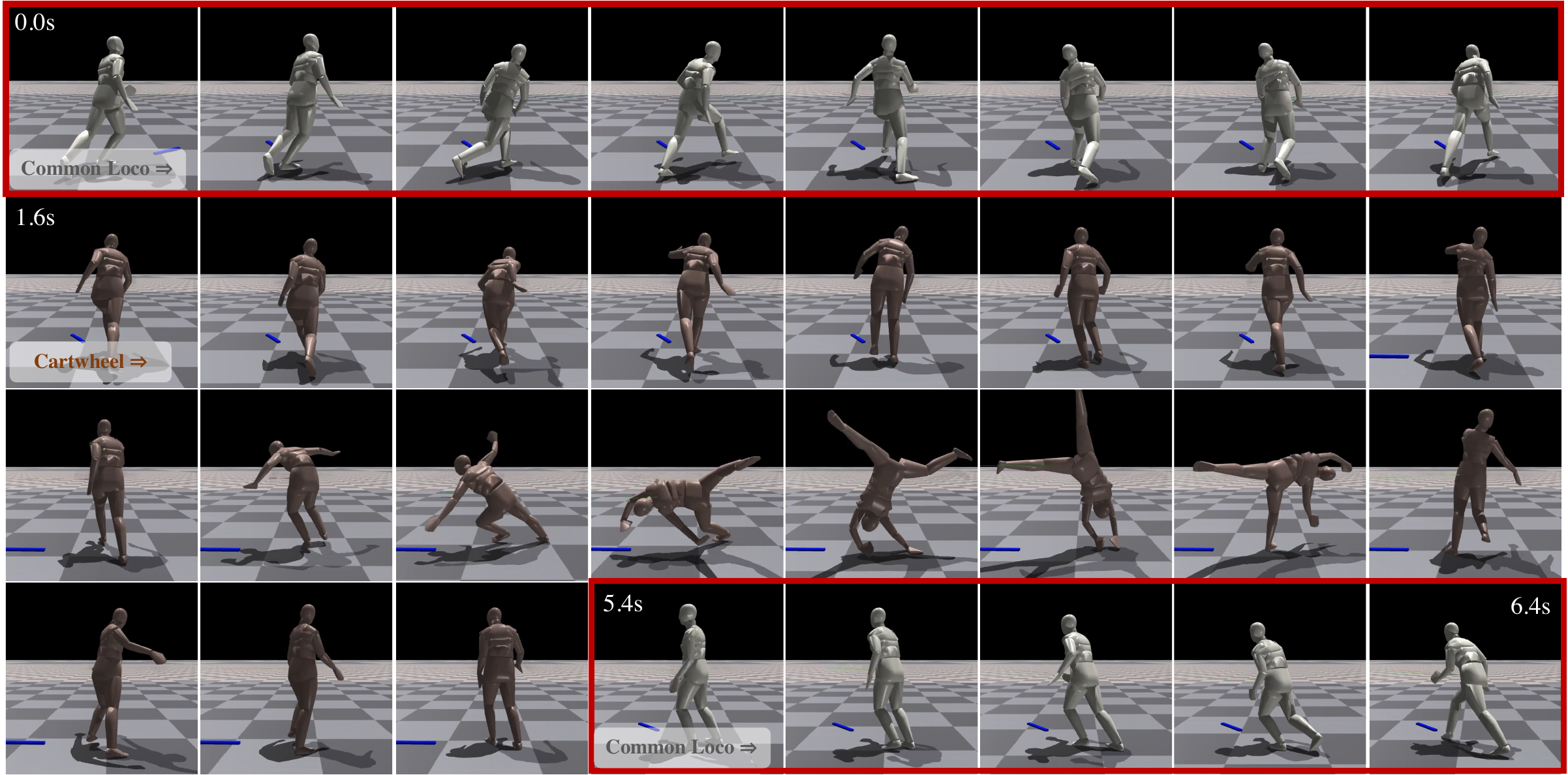}
    \vspace{-2mm}
    \caption{\textbf{Example results of locomotion control with skills of Cartwheel and Common Locomotion} shown in 5 fps. The gamepad control signal is marked with a blue stick. The color of the simulated character marks different retrieval databases $\{\mathcal{M}^\text{retr}_k|k\}$ chosen by users at run time. Light Grey: Common Locomotion. Brown: Cartwheel.}
    \vspace{-4mm}
    \label{fig:runcartwheel}
\end{figure}

\noindent\textbf{Cartwheel and Common Locomotion Qualitative Results.}
Controlling a simulated character in stable and balanced body states such as walk\&run is comparatively straightforward since the center of mass (CoM) is typically positioned between the feet. In contrast, performing a 'cartwheel' presents a more complex locomotion challenge.

Our results, demonstrating the transition between common locomotion and a cartwheel, are presented in Fig.~\ref{fig:runcartwheel}. Initially, the character is guided by the reference retrieved from the common locomotion database. We then shift to the cartwheel database in $\mathbf{Env}^\text{retr}$ by simply switching the lookup dictionary, indicated by the color of brown. We provide further qualitative results in the supplementary video.

Without any additional planning required, our motion switching performance at run-time showcasing the flexibility and adaptability of our system. 
Notably, our proposed approach showcases the extraordinary versatility of even a lightweight MLP controller, which can cover a diverse range of distinctive modes. Additionally, the simulated character is able to execute specific styles while simultaneously reaching the user-designated goal, further underscoring the responsiveness of our approach.

\begin{table}
    \centering
    \caption{\textbf{Quantitative comparison on the basic locomotion.}}
    \setlength{\tabcolsep}{1mm}
    \resizebox{\columnwidth}{!}{
    \begin{tabular}{c|ccccc@{}}
    \bottomrule \hline
        \makecell[c]{Method} & \makecell[c]{MVE(m/s) $\downarrow$} & \makecell[c]{TRate(\%) $\downarrow$} & \makecell[c]{Len(\%) $\uparrow$} & \makecell[c]{FID $\downarrow$} & \makecell[c]{MModality $\nearrow$}\\
    \hline
    \makecell[c]{Real Motion} & \makecell[c]{-} & \makecell[c]{-} & \makecell[c]{-} & \makecell[c]{0.0113} & \makecell[c]{1.0271} \\ 
    \hline
        \makecell[c]{DReCon (TOG'19)} & \makecell[c]{3.361} & \makecell[c]{28.06} & \makecell[c]{82.98} & \makecell[c]{7.998} & \makecell[c]{\textbf{3.386}} \\ 
        \makecell[c]{AMP (TOG'21)} & \makecell[c]{3.551} & \makecell[c]{5.00} & \makecell[c]{96.01} & \makecell[c]{\textbf{3.453}} & \makecell[c]{1.096} \\ 
        \makecell[c]{ASE (TOG'22)} & \makecell[c]{3.472} & \makecell[c]{1.65} & \makecell[c]{98.52} & \makecell[c]{4.212} & \makecell[c]{0.727} \\
    \hline
        \makecell[c]{Ours} & \makecell[c]{\textbf{2.709}} & \makecell[c]{\textbf{0.44}} & \makecell[c]{\textbf{99.69}} & \makecell[c]{\textbf{3.453}} & \makecell[c]{1.380} \\ 
        \makecell[c]{Ours (${s}^{\text{retr}}$)} & \makecell[c]{2.837} & \makecell[c]{-} & \makecell[c]{-} & \makecell[c]{-} & \makecell[c]{-} \\ 
    \hline \toprule  
    \end{tabular}
    }
    \vspace{-3mm}
    \label{tab:StatCompare1}
\end{table}

\noindent\textbf{Common Locomotion Quantitative Comparisons.}
We show quantitative results of common locomotion control in Table~\ref{tab:StatCompare1}, compared with DReCon~\cite{bergamin2019drecon}, AMP~\cite{peng2021amp} and ASE~\cite{peng2022ase}. 

We employ various evaluation metrics including Mean Velocity Error (MVE), Termination Rate (TRate), Episode Length (Len), Fréchet Inception Distance (FID), and Multimodality (MModality). We provide an explanation of these metrics in the supplement.

The robustness of our proposed system is evidenced by the lowest TRate and highest Len. This accomplishment is partly due to the similar GAIL training scheme used by AMP, ASE, and our approach, which significantly surpasse DReCon.
\begin{figure}[ptb]
    \centering
    \includegraphics[width=0.9\columnwidth]{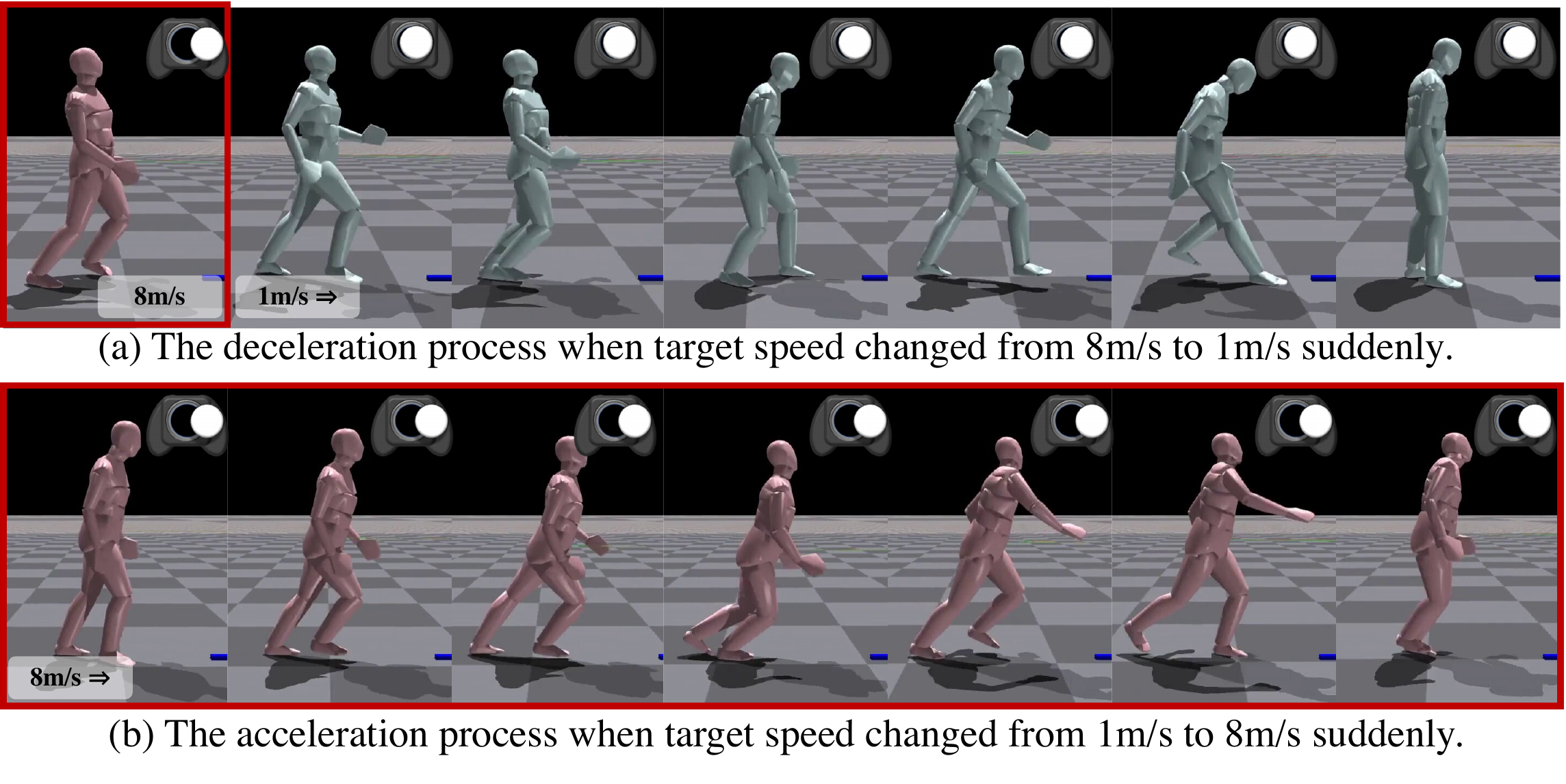}
    \vspace{-2mm}
    \caption{\textbf{Generalization ability and response latency of our system.} We set an extreme case of high and low speed to 8m/s and 1m/s. The simulated character in red indicates that it is currently under the goal with a higher speed, while it switches to a lower speed when turning to green.}
    \vspace{-4mm}
    \label{fig:accdeacc}
\end{figure}
Furthermore, our method's use of the task-oriented retriever allows it to achieve the smallest MVE amongst the comparisons, suggesting that our proposal efficiently responds to the locomotion control signal. We evaluated intermediate retrieved states (${s}^{\text{retr}}$) from our method, revealing a small MVE, which shows an automatic improvement in retrieval performance achieved by task-oriented optimization. Importantly, our reference mimic objective isn't tailored to exactly replicate the expert motion but instead to provide end effector information, thus enabling the system (Ours) to outperform the retrieved expert (Ours ${s}^{\text{retr}}$). The substantial absolute value of MVE can be attributed to latency stemming from the natural acceleration and deceleration process, as depicted in Fig.~\ref{fig:accdeacc}.

In terms of motion diversity and fidelity, a comprehensive assessment of MultiModality performance necessitates the consideration of the FID metric as well. The generated motion is expected to closely resemble real motion while demonstrating a comparable diversity. While DReCon can control motion with diverse appearance, it often results in unnatural outcomes with high FID.

\begin{figure}[ht]
    \centering
    \includegraphics[width=0.7\columnwidth]{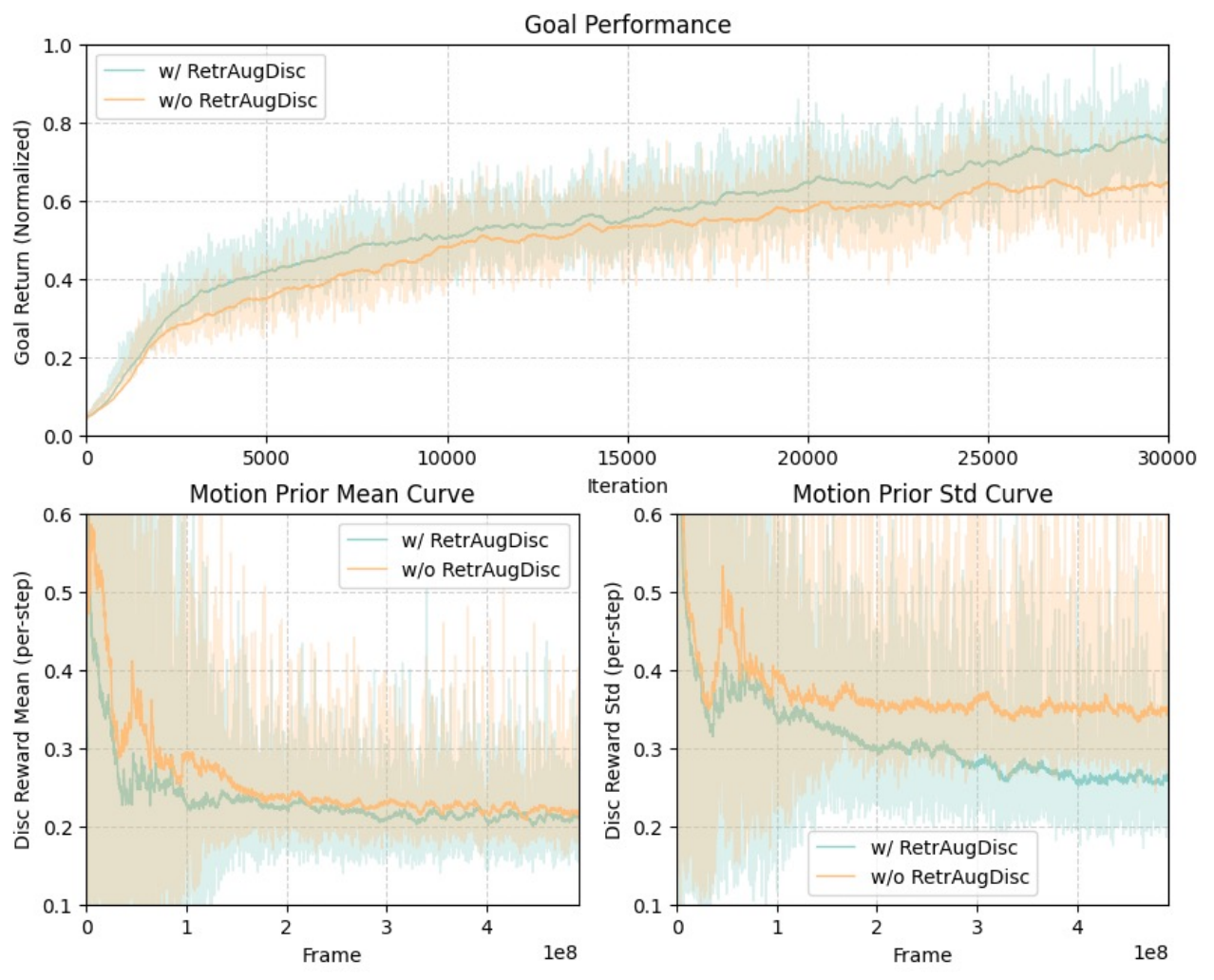}
    \vspace{-2mm}
    \caption{\textbf{Ablation study.} We show the training curves of policies learned on a simple locomotion task w/ and w/o retrieval augmentation (RetrAugDisc). The Goal Performance shows the normalized overall goal return of an episode.}
    \vspace{-4mm}
    \label{fig:ablation}
\end{figure}

\begin{table}[ht]
    \centering
    \vspace{-4mm}
    \caption{\textbf{Effect of Modules.} Compared in terms of Mean Velocity Error (MVE, m/s), Terminate Rate (TRate, \%), Episode Length (Len, \%), Frechet Inception Distance (FID), and Multimodality (MModality).}
    \setlength{\tabcolsep}{1mm}
    \resizebox{\columnwidth}{!}{
    \begin{tabular}{c|ccccc@{}}
    \bottomrule
        \makecell[c]{Method} & \makecell[c]{MVE(m/s) $\downarrow$} & \makecell[c]{TRate(\%) $\downarrow$} & \makecell[c]{Len(\%) $\uparrow$} & \makecell[c]{FID $\downarrow$} & \makecell[c]{MModality $\nearrow$}\\
    \hline
        \makecell[c]{Ours(${s}^{\text{retr}}$)} & \makecell[c]{2.837} & \makecell[c]{-} & \makecell[c]{-} & \makecell[c]{-} & \makecell[c]{-} \\
        \makecell[c]{w/o learnable retriever} & 
        \makecell[c]{\textbf{2.387}} & \makecell[c]{-} & \makecell[c]{-} & \makecell[c]{-} & \makecell[c]{-} \\ 
    \hline
        \makecell[c]{Ours} & \makecell[c]{2.709} & \makecell[c]{\textbf{0.44}} & \makecell[c]{\textbf{99.69}} & \makecell[c]{\textbf{3.453}} & \makecell[c]{1.380} \\ 
        \makecell[c]{w/o learnable retriever} & \makecell[c]{3.170} & \makecell[c]{8.71} & \makecell[c]{91.54} & \makecell[c]{4.178} & \makecell[c]{1.593} \\ 
        \makecell[c]{w/o RA-GAIL} & \makecell[c]{2.610} & \makecell[c]{4.37} & \makecell[c]{95.37} & \makecell[c]{3.770} & \makecell[c]{\textbf{1.716}} \\
    \bottomrule  
    \end{tabular}
    }
    \vspace{-3mm}
    \label{tab:Statablation}
\end{table}

\noindent\textbf{Ablation Study}
We provide further evidence of the strength of the retrieval-augmented adversarial prior, demonstrating its ability to promote more stable training. Fig.~\ref{fig:ablation} presents the training curves of policies learned on a straightforward locomotion task (using the common locomotion database only), comparing those with and without RetrAugDisc.
For the sake of consistency and simplicity in our experiments, we employ a standard unlearnable retriever to query the retrieved expert clips. The policy trained with a retrieval-augmented adversarial prior not only learns faster, but it also shows a higher goal return in the end.
Our policy, trained with a retrieval-augmented adversarial prior, yields a smaller training reward variance, indicating a more stable training process. We also report the quantitative results to demonstrate the efficacy of our end-to-end learnable retriever and RA-GAIL in~\cref{tab:Statablation}.

\section{Conclusion}

In conclusion, we have constructed an end-to-end system with a task-oriented learnable retriever and a simulated character controller, culminating in a retrieval-augmented hierarchical reinforcement learning system for physics-based humanoid locomotion control. This system is not only interpretable, but also demonstrates exceptional generalizability across diverse motion types. This advancement has the potential to positively influence a broad spectrum of applications, including virtual reality, animation, and more.

\bibliographystyle{IEEEbib}
\small
\bibliography{strings,refs}

\end{document}